\documentclass[10pt]{article} % For LaTeX2e
\usepackage[preprint]{tmlr}
\usepackage{amsmath,amsfonts,bm}

\def\eqref#1{equation~\ref{#1}}
\def\1{\bm{1}}

\DeclareMathAlphabet{\mathsfit}{\encodingdefault}{\sfdefault}{m}{sl}
\SetMathAlphabet{\mathsfit}{bold}{\encodingdefault}{\sfdefault}{bx}{n}

\usepackage{hyperref}
\usepackage{url}
\usepackage{graphicx}
\usepackage{array}
\usepackage{booktabs}
\usepackage{tabularx}
\usepackage{pifont}
\usepackage{multirow}

\title{Early Quantization Shrinks Codebook: \\ A Simple Fix for Diversity-Preserving Tokenization}

\author{\name Wenhao Zhao\thanks{Equal Contribution} \email e1374536@u.nus.edu \\ 
\addr National University of Singapore \AND
\name Qiran Zou\footnotemark[1] \email qiranzou@gmail.com \\ 
\addr National University of Singapore \AND
\name Rushi Shah \email shah.15@iitj.ac.in \\ 
\addr National University of Singapore \AND
\name Yudi Wu \email wuyudi@zju.edu.cn \\
\addr Zhejiang University \AND
\name Zhouhan Lin \email lin.zhouhan@gmail.com \\
\addr Shanghai Jiao Tong University \AND
\name Dianbo Liu \email dianbo@nus.edu.sg \\
\addr National University of Singapore
}

\def\month{MM}  % Insert correct month for camera-ready version
\def\year{YYYY} % Insert correct year for camera-ready version
\def\openreview{\url{https://openreview.net/forum?id=XXXX}} % Insert correct link to OpenReview for camera-ready version

\begin{document}

\maketitle

\begin{abstract}

While discrete tokenizers are suspected to inherently limit sample diversity in token-based generative models, we show this diversity gap is not caused by discretization itself, but rooted in the timing of quantization. 
In this study, we systematically identify quantization in the initial stage as the primary catalyst for a representational misalignment, where the codebook prematurely shrinks into a narrow latent manifold. This initial shrinkage prevents the codebook from capturing the diverse embedding space of the encoder.
% Relation with  generative diversity
Though this may yield deceptively strong reconstructions, it creates a bottleneck that forces the generator to rely on a homogenized set of tokens. Ultimately, the codebook’s failure to anchor to robust representations at the onset of training impairs generative variety and limits sample diversity.
% How to solve it?
To address this, we propose Deferred Quantization, a simple yet effective strategy that introduces a separate continuous learning phase. By allowing the encoder to first establish a well-distributed representation space before introducing discretization, the codebook can effectively anchor to a mature and diverse latent landscape.
% show what contribution we make
Across tokenizers and token-based generators, Deferred Quantization consistently reduces shrinkage, improves generative diversity, and preserves reconstruction and compression. We additionally provide a shrinkage diagnostic suite and offer practical guidance for designing diversity-preserving discrete tokenizers.
\end{abstract}

\section{Introduction}
\begin{figure*}[ht]
    \centering
    \includegraphics[width=0.9\linewidth]{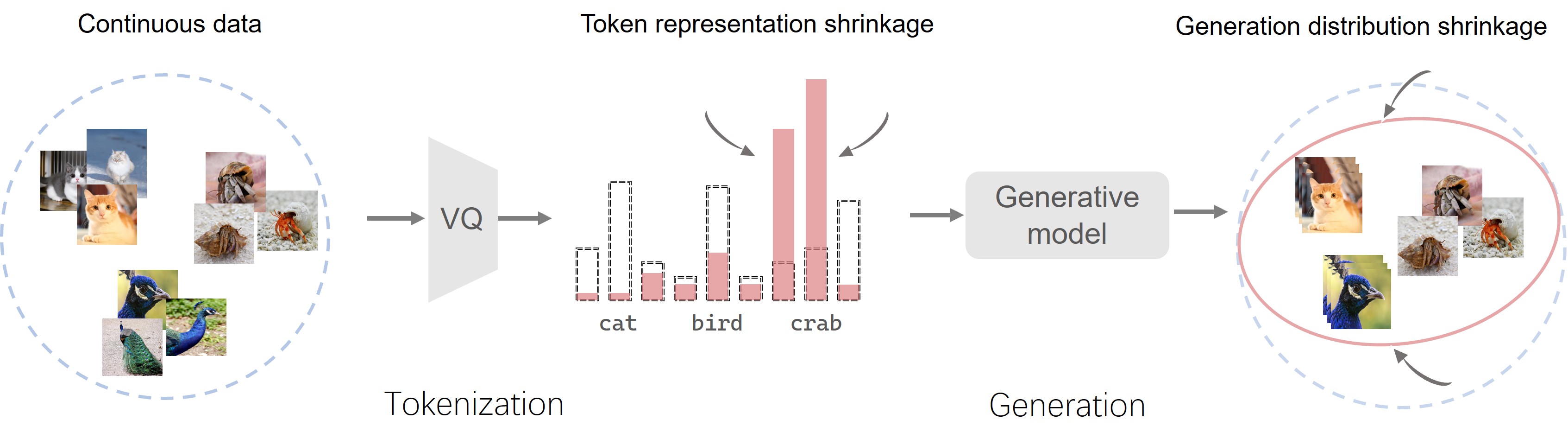}
    \caption{\textbf{Token representation shrinkage degrade the diversity of generative model.} 
    Vector quantization is a widely used technique to map continuous data into discrete tokens, which assists the transformer-based generative model's generation. 
    We observe that token representation shrinkage, manifested as narrow distribution in latent space, leads to a shrunk distribution of the generated data. 
    }
\label{fig:token_shrinkage_illustration}
\end{figure*}
\vspace{-0.5em}

Transformer-based generative models for autoregressive generation have gained significant popularity in recent years in the field of image generation. These models underpin many state-of-the-art systems such as DALL-E \citep{betker2023improving} and VAR \citep{tian2024visual}, which have found wide-ranging applications in art creation, design automation, and data augmentation. Their practical value lies not only in producing visually compelling images but also in enabling new workflows for creative and industrial domains.

Despite their success, transformer-based generative models may exhibit a limitation: the synthetic images they produce tend to cover a narrower distribution than real images. This phenomenon, commonly referred to as mode collapse, results in limited diversity in the generated content. Mode collapse causes the model to ignore valid variations in the data distribution, which limits its generalization, realism, and utility in downstream tasks. In this work, we study on the ability of generative models to produce diverse high-quality outputs from the perspective of tokenization, and study how limitations in diversity arise in token-based generation pipelines.

A common implicit assumption in token-based generation is that a tokenizer is good if it yields low reconstruction error. However, reconstruction primarily reflects performance on the high-density regions of the data distribution, while diversity depends critically on broader coverage of the representation space, including low-density regions. This mismatch suggests that a tokenizer can appear strong under reconstruction-centric evaluation, yet still bottleneck downstream diversity.

To address mode collapse and improve generative fidelity and diversity, various studies have proposed architectural innovations or alternative training objectives. For example, VQGAN \citep{esser2021taming} incorporates vector quantization to learn a discrete codebook, while ImageGPT \citep{chen2020generative} treats images as sequences of pixels to better capture complex data distributions. However, these advances do not directly characterize whether the learned discrete tokens sufficiently cover the latent representation space required for diverse generation.

In this work, we identify a previously overlooked yet critical aspect of tokenization, which we term \textit{token representation shrinkage}. It describes a phenomenon in which the learned token embeddings shrink into a small region of the latent space, causing most tokens to cluster around only a few modes, as illustrated in Fig.~\ref{fig:token_shrinkage_illustration}. When tokens shrink into a limited region of the distribution, the generated outputs are also constrained to a narrow portion of the data space, resulting in reduced diversity and diminished modality coverage. Therefore, token representation shrinkage provides a concrete mechanism by which tokenization can limit downstream diversity even when reconstruction appears satisfactory.

Importantly, existing studies \citep{zhu2025addressing,yu2021vector,yu2023language,mentzer2023finitescalarquantizationvqvae} on codebook collapse have mainly focused on the dead-token problem, where some codebook entries are never or rarely used during training, while largely overlooking the issue of insufficient coverage of the embedding space, \emph{i.e.}, token representation shrinkage. This distinction matters because a codebook may exhibit healthy usage statistics while still occupying only a small region of the latent space, thereby restricting the support available to downstream generative models.

To better understand this issue, we analyze its underlying cause and identify a key contributing factor: the commonly used token initialization strategy \citep{zeghidour2021soundstream} in VQ training. In typical practice, token embeddings are initialized from the outputs of an untrained encoder, which already form a compact and clustered distribution in the latent space. This initialization bias restricts the subsequent expansion of the token space during training, making it difficult for the learned tokens to spread out and align with the true data distribution. As a result, it induces token representation shrinkage.

We then introduce a simple intervention Deferred Quantization that alleviates token representation shrinkage and enables a controlled study of its impact: we first pretrain the encoder without VQ and then utilize semantic embeddings to initialize the codebook. This decouples representation learning from discretization in the early stage, allowing the encoder to learn meaningful semantic representations before quantization is introduced. As a result, it reduces the resistance faced during VQ optimization and mitigates the shrinkage effect. Using this intervention, we validate the presence and impact of token representation shrinkage through extensive experiments on a variety of synthetic and real-world datasets. Across these settings, we observe that token representation shrinkage is associated with reduced diversity, whereas mitigating shrinkage reliably improves both the diversity and fidelity of generated images.

Our main contributions are summarized as follows:

\begin{itemize}
    \item We analyze the detrimental effects of early quantization, demonstrating it causes the codebook to shrink into a narrow, homogenized latent space, which consequently impairs generative sample diversity
    
    \vspace{-0.25em}
    \item We introduce Deferred Quantization, a simple yet effective training intervention that decouples feature learning from discretization. 

    \vspace{-0.25em}
    \item We validate this Deferred Quantization across synthetic and real-world datasets, showing that mitigating token representation shrinkage consistently restores the codebook's representational breadth and improves generative variety.
    
\end{itemize}

\section{Related Works}
Vector quantization plays an important role in data compression and signal processing under Shannon’s rate–distortion theory \citep{gersho2012vector, cover1999elements}. Traditional approaches such as K-means clustering \citep{mcqueen1967some} have been widely used, but they become computationally expensive when applied to high-dimensional data \citep{le2018deepvq}.
To mitigate this challenge, DeepVQ \citep{le2018deepvq} improved efficiency by mapping data to lower-dimensional latent spaces before quantization. 
Moreover, \citep{van2017neural} proposed VQ-VAE which integrates VQ with variational autoencoders, using a straight-through estimator \citep{bengio2013estimating} to handle discrete variables. 
To refine VQ methods for improved performance, variants such as Residual Quantization \citep{lee2022autoregressive}, Product Quantization \citep{chen2020differentiable}, and Soft Convex Quantization \citep{gautam2023soft} further enhanced representation capacity and efficiency. Recent advances incorporate attention mechanisms and transformer architectures \citep{2017Attention, yu2021vector} to dynamically select codebooks and capture global data dependencies.
Recent works also explore per-channel codebooks \citep{hsu2024disentanglement} and neural network variants of residual quantization \citep{2024residual} to predict specialized codebooks, enhancing the model's expressive power.

% --------------- Application of VQ ----------------- %
VQ has been widely applied across various domains. In natural language processing, VQ facilitates sequence modeling \citep{kaiser2018fast}, enhancing tasks such as language modeling. %and machine translation.
In computer vision, VQ has significantly advanced image generation and compression techniques \citep{esser2021taming}. 
Similarly, in audio processing, VQ techniques have captured complex temporal dependencies \citep{dhariwal2020jukebox}. 
Furthermore, in multimodal applications, VQ supports the integration of different data types through shared discrete representations \citep{ramesh2021zero}.

% ---------------- Limitation/Problems in VQ ----------- %
Despite these advancements, VQ methods encounter challenges that restrict their broader application, including but not limited to codebook collapse, training instability, and computational overhead. 
Extensive research has been conducted on solving the codebook collapse problem,  where only a subset of tokens are used leading to inefficient representation usage and reduced diversity in outputs, by reducing token dimension \citep{yu2021vector}, orthogonal regularization loss \citep{shin2023exploration}, multi-headed VQ \citep{mama2021nwt},  finite scalar quantization \citep{mentzer2023finitescalarquantizationvqvae}, and Lookup Free Quantization \citep{yu2023magvit}.
Recent methods like \citep{goswami2024hypervq} and \citep{baykal2024edvae} also strive to enhance tokens usage efficiency.
However, beyond the widely recognized issue of codebook collapse, the token representation shrinkage we identify has not been previously recognized in the literature.

\section{Preliminary}

\subsection{Mode Collapse and Limited Diversity}
Transformer-based generative models might exhibit a limitation: the synthetic images they produce tend to cover a narrower distribution than real images. This phenomenon, commonly referred to as \emph{mode collapse}, results in limited diversity in the generated content. Mode collapse causes the model to ignore valid variations in the data distribution, which limits its generalization, realism, and utility in downstream tasks.

In this paper, we use \emph{mode collapse} to refer to the reduction in diversity and modality coverage of generated outputs relative to the data distribution. In particular, mode collapse manifests when the generative model concentrates probability mass on a small subset of plausible outcomes, underrepresenting other valid modes of variation in the data. As a result, mode collapse and limited diversity undermine the effectiveness of generative models in settings that require broad coverage of the data distribution.

\subsection{Vector Quatization}

\paragraph{VQ-VAE} We define the VQ-VAE as following: an encoder $E_\theta$, a decoder $D_\theta$ and a set of tokens $\mathcal{T}=\{t_1, t_2,\ldots, t_S\}$.  The token set $\mathcal{T}$ constitutes the codebook, which is employed to store the discretized representations. The encoder is responsible for mapping the raw data $X=\{x_1, x_2, \ldots, x_N \}$ to a set of continuous representations $\mathcal{Z}=E_\theta(X)$, 
where $\mathcal{Z}=\{z_1, z_2, \ldots, z_N \}$.
And the decoder reconstructs the data $X'=D_\theta(\hat{Z})$
based on the set of discretized representations $\hat{Z}$, where $\hat{Z}=\{\hat{z}_1, \hat{z}_2, \ldots, \hat{z}_N \}$.The process of tokenizing a continuous representation $z_j$ to discrete representation $\hat{z_j}$ is as following:
    \begin{equation}
        \hat{z_j} = \arg\min_{t_k \in \mathcal{T}} \| z_j - t_k \|,
    \label{equ_discretization_process}
    \end{equation}
where $t_k$ is a token in token set $\mathcal{T}$ and $k$ is the index. This quantization is performed by finding the nearest token $t_k$ in $\mathcal{T}$.The optimization objective comprises reconstruction loss $\mathcal{L}_{\text{recon}}$, codebook loss $\mathcal{L}_{\text{codebook}}$, and commitment loss $\mathcal{L}_{\text{commit}}$.

\section{Token Representation Shrinkage}
\label{sec:Shrinkage Phenomena and Sythentic Experiments Results}

\begin{figure}[t]
    \centering
    \includegraphics[width=1.0\linewidth, clip, trim={0.8cm, 0.4cm, 0.0cm, 0.0cm}]{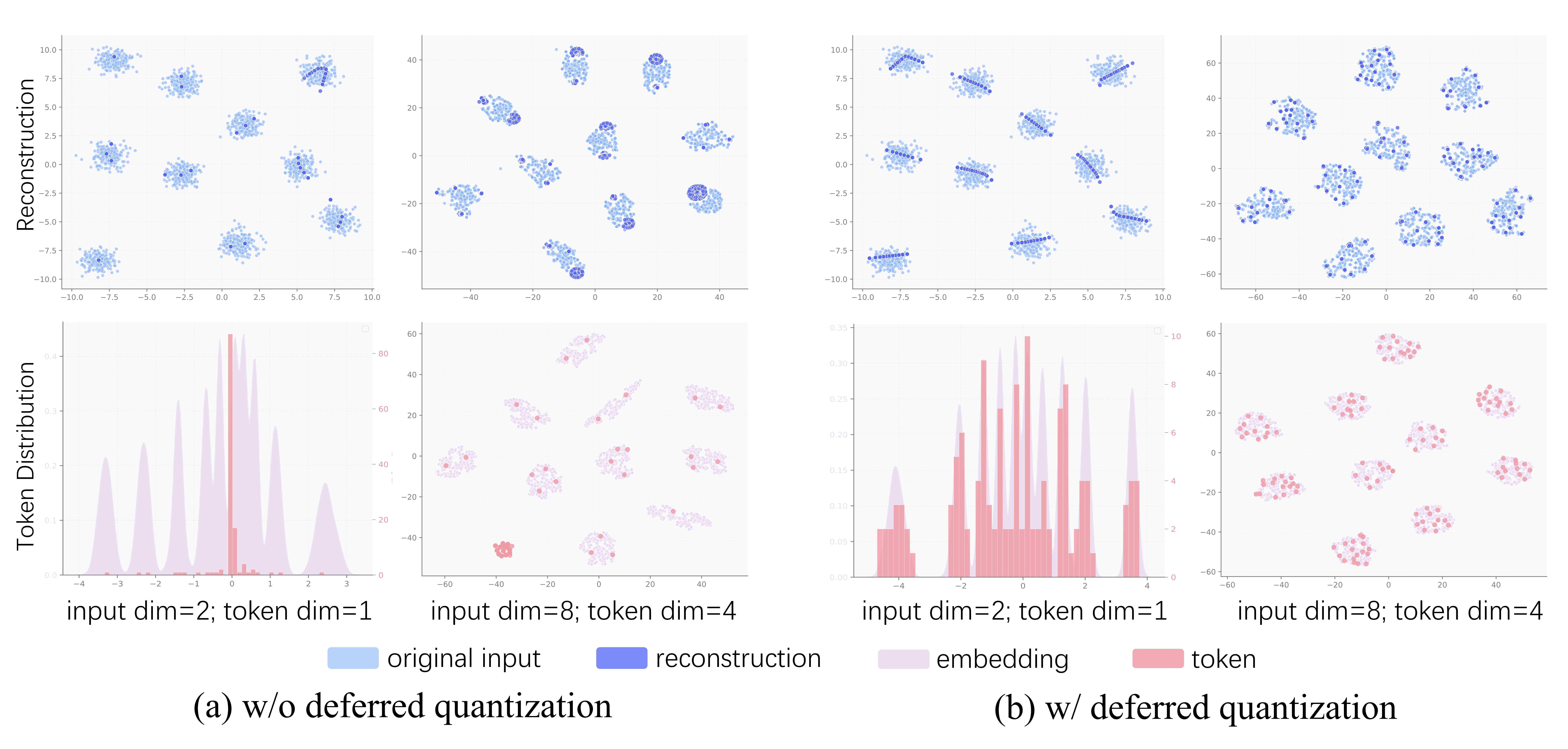}
    \caption{
        \textbf{Token representation shrinkage shrinks latent support and induces reconstruction mode collapse while Deferred Quantization mitigates it.}
        (a) \textbf{w/o deferred quantization:} tokens cluster into a narrow region of the embedding space, reducing coverage and causing reconstructions to collapse onto fewer modes across different input dimensions.
        (b) \textbf{w/ deferred quantization:} tokens spread across the embedding space, improving latent support coverage and yielding reconstructions with better modality coverage.
    }
    \label{fig:tokens_collapse_syn_all_exp_results}
\end{figure}

This section characterizes \emph{token representation shrinkage} and analyzes why it can induce downstream mode collapse and reduced diversity in token-based generation pipelines.

\subsection{Empirical Characterization}
\label{sec:shrinkage_empirical}

\paragraph{Definition.}
Token representation shrinkage refers to a geometric failure mode where learned token embeddings occupy only a small region of the encoder embedding space, causing many tokens to cluster around a few modes, as illustrated in Fig.~\ref{fig:token_shrinkage_illustration}. In the ideal case, the token distribution should align with and adequately cover the encoder embedding distribution; shrinkage violates this coverage requirement.

\paragraph{Synthetic evidence: shrinkage induces reconstruction collapse.}
To validate the phenomenon, we conduct experiments on a synthetic dataset using VQ-VAE. Specifically, we use VQ-VAE to reconstruct the input data and compare the resulting token distribution with the original data distribution. The synthetic dataset comprises 10,000 data points uniformly sampled from 10 distinct Gaussian distributions (see Sec.~\ref{sec:experiment_setup} for details).

As shown in Fig.~\ref{fig:tokens_collapse_syn_all_exp_results} (a), the learned tokens densely cluster within a limited region of the latent space, and the reconstructed data distribution collapses accordingly. Consequently, reconstructions fail to capture the full modality spectrum of the original data, exhibiting a clear form of mode collapse at the representation level.

\begin{figure}[t]
    \centering
    \includegraphics[width=0.75
    \linewidth]{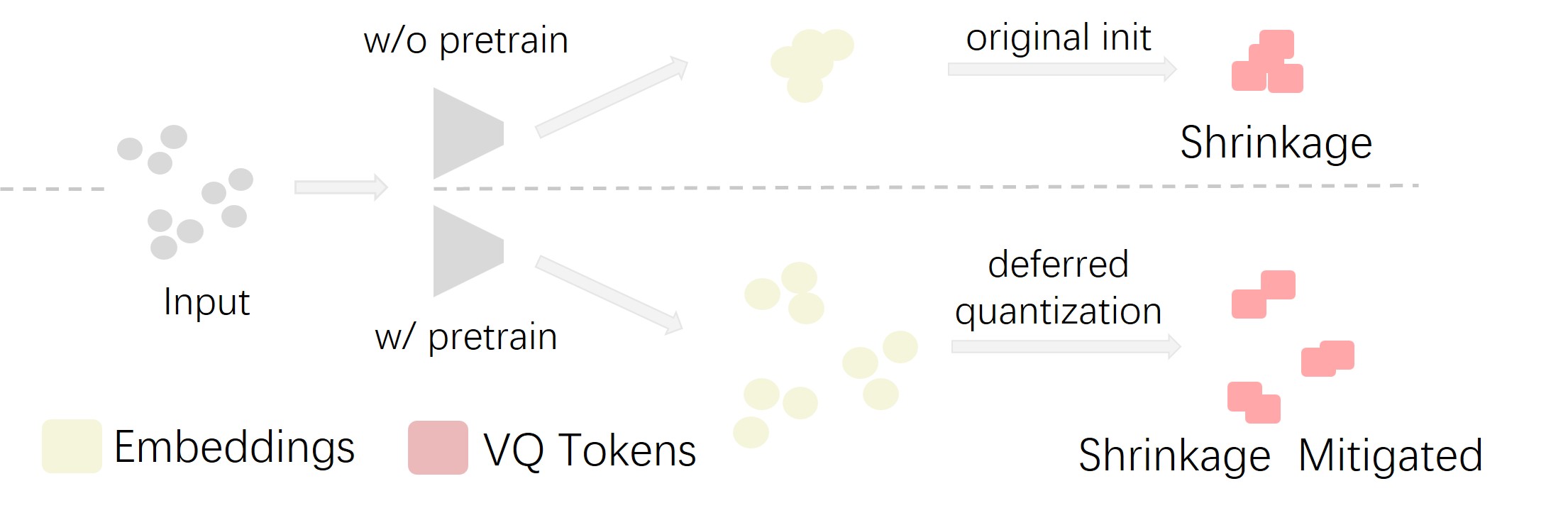}
    \caption{\textbf{Early quantization with clustered initialization induces token representation shrinkage while Deferred Quantization mitigates it.}
    Initializing the codebook from an untrained encoder yields a narrow, uninformative embedding distribution, causing tokens to cluster and shrink latent support at early stage. In contrast, Deferred Quantization first learns a dispersed continuous representation and then initializes the codebook with semantic embeddings from the pretrained encoder, yielding better token coverage and reducing shrinkage.}
\label{fig:token_shrinkage_and_pretrained_methods}
\end{figure}

\paragraph{Root cause for shrinkage: clustered initialization from an untrained encoder.}
A key contributing factor is the common codebook initialization practice: token embeddings are initialized from the outputs of an untrained encoder, whose embeddings are already concentrated in a narrow region of the latent space. As shown in Fig.~\ref{fig:token_shrinkage_and_pretrained_methods}, the output distribution of an untrained encoder is significantly more concentrated than that of a trained encoder.

To examine how this initialization contributes to shrinkage, we compare embedding distributions produced by trained and untrained encoders on the synthetic dataset. We observe that the untrained encoder yields embeddings with fewer distinct peaks and a narrower spread, suggesting fewer distinguishable modes. Since an untrained encoder lacks meaningful feature extraction ability, it maps diverse inputs to similar embeddings, leading to clustered token initialization and insufficient coverage of the embedding space. Further experimental details and visualizations are provided in the appendix \ref{appendix_Implementation_details}.

\paragraph{Intervention for controlled study.}
Building on these observations, we hypothesize that initializing tokens using an encoder that has already learned semantic distinctions---and therefore produces dispersed embeddings---should improve token coverage and mitigate shrinkage. This motivates a simple intervention: \emph{train the autoencoder without VQ first, then initialize the codebook using semantic embeddings from the pretrained encoder}. Concretely, we (i) train an autoencoder, then (ii) train a VQ-VAE initialized with the autoencoder weights.

Pretraining allows the encoder to learn a more structured and spread representation space, providing a more informative foundation for codebook initialization, as shown in Fig.~\ref{fig:token_shrinkage_and_pretrained_methods}.

\paragraph{Result: mitigating shrinkage improves modality coverage.}
We evaluate this intervention on the synthetic dataset (Fig.~\ref{fig:tokens_collapse_syn_all_exp_results}). Comparing subfigures (a) and (b), mitigating token representation shrinkage yields a more uniform token distribution and a reconstruction distribution that aligns more closely with the original input distribution. Notably, under higher input dimensions (input dim=8), shrinkage leads to substantially degraded reconstruction and collapsed modality coverage, whereas the mitigated version remains closer to the ground-truth distribution. These results support the view that token representation shrinkage is a concrete mechanism that can induce mode collapse and reduced diversity.

\subsection{Theoretical Analysis}
\label{sec:shrinkage_theory}

We provide a simplified analysis to formalize how token representation shrinkage can reduce diversity and increase distortion.

\paragraph{Setup.}
Consider a data distribution constructed from \(K\) well-separated and equally weighted component distributions \(p(x|k)\),
\begin{equation}
p(x) = \sum_{k=1}^{K} p(x|k)p(k) = \frac{1}{K}\sum_{k=1}^{K}p(x|k),
\label{equ_dataset}
\end{equation}
where \(p(k)=\frac{1}{K}\).
For simplicity, assume the encoder and decoder are identity mappings (i.e., \(x'=\mathrm{Dec}(\mathrm{Enc}(x))=x\)), and that the transformer can perfectly model the token distribution. Under these assumptions, the only source of distortion arises from vector quantization. We measure distortion by the expected squared error:
\begin{equation}
\mathcal{E} = \mathbb{E}_{x \sim p} \left[ \|q(x) - x\|_2^2 \right],
\label{equ_quality}
\end{equation}
where \(q\) is the quantization function.

\paragraph{Diversity proxy.}
We use the entropy \(H\) of the induced mode distribution as a proxy for diversity:
\begin{equation}
H = -\sum_{k=1}^{K} p_k \log p_k,
\label{equ_diversity}
\end{equation}
with
\begin{equation}
p_k = \frac{|T_k|}{\sum_{j=1}^{K} |T_j|}, \qquad
T_k = \{\, t_i \mid t_i \in \text{cluster } k \,\},
\label{equ_pk_tk}
\end{equation}
where \(T_k\) is the set of tokens assigned to cluster \(k\), and \(p_k\) is the empirical probability mass on that cluster.

\paragraph{Effect of shrinkage.}
In the ideal balanced case, \(p_k \approx 1/K\), yielding maximal entropy \(H=\log K\) and lower distortion. Under token representation shrinkage, token mass concentrates on a subset of modes \(\mathcal{J}\subset\{1,\ldots,K\}\) with \(|\mathcal{J}|=M\ll K\). In this case, entropy is upper bounded by
\begin{equation}
H \le \log M,
\label{equ_entropy_shrink}
\end{equation}
implying a strict reduction in diversity compared to the balanced optimum \( \log K \).
Moreover, samples from inactive modes \(k\notin \mathcal{J}\) are forced to encode using distant tokens, increasing quantization error and therefore increasing \(\mathcal{E}\). Together, these effects formalize how shrinkage can simultaneously reduce diversity and increase distortion, which in turn can manifest as mode collapse in downstream token-based generation.

\section{Deferred Quantization for Diversity-Preserving Tokenization}
\label{sec:deferred_quantization}

This section introduces \emph{Deferred Quantization}, a simple training protocol that decouples early representation learning from discretization to mitigate token representation shrinkage.

\paragraph{Motivation}
Token representation shrinkage is strongly influenced by the geometry of the encoder embedding space at the time quantization is enabled. When quantization is applied before the encoder learns a sufficiently spread representation, nearest-center assignment can reinforce shrinkage, leading to insufficient coverage.

\paragraph{Two-stage training protocol}
We adopt a two-stage procedure:
\textbf{Geometric Phase (continuous training).} Train an autoencoder without vector quantization, allowing the encoder to learn a structured, dispersed embedding space.
\textbf{Discretization Phase (enable VQ).} Initialize the codebook using semantic embeddings from the pretrained encoder, then enable vector quantization and continue training under the VQ objective.
This protocol encourages the codebook to adapt to an already-formed representation manifold, mitigating shrinkage and improving coverage.

Deferred Quantization is intentionally minimal: it does not change the generator architecture and introduces no additional inference cost. It directly targets a training-time mechanism (insufficient early coverage) that leads to token representation shrinkage, thereby improving diversity and reducing mode collapse in token-based generation pipelines.

\section{Experiments}

In this section, we first validate that token representation shrinkage arises in real-world datasets (CIFAR-10 and ImageNet-100), and then show that shrinkage can propagate to downstream token-based generators as reduced sample diversity and worsened mode coverage (often manifesting as mode collapse). We conduct experiments on two representative token-based generative models, MaskGIT \citep{chang2022maskgit} and VAR \citep{tian2024visual}, using both ImageNet-100 and a medical image dataset. To better elucidate the logical connections between our claims and empirical findings, we provide a claim-evidence map in Tab~\ref{tab:claim_evidence_shrinkage}.

\begin{table*}[t]
\centering
\scriptsize
\setlength{\tabcolsep}{5pt}
\renewcommand{\arraystretch}{1.2}
\vspace{-0.5em}
\caption{\textbf{Claim-Evidence map for token representation shrinkage and Deferred Quantization.} This table summarizes our key theoretical claims and maps them to specific empirical evidence, visualizations, and quantitative metrics provided in the paper.}
\vspace{1.0em}
\label{tab:claim_evidence_shrinkage}
\begin{tabularx}{\textwidth}{l >{\raggedright\arraybackslash}p{0.32\linewidth} >{\raggedright\arraybackslash}X >{\raggedright\arraybackslash}p{0.26\linewidth}}
\toprule
\textbf{ID} & \textbf{Claim} & \textbf{Evidence} & \textbf{Metrics} \\ 
\midrule

C1 & Root cause of shrinkage is clustered codebook initialization from untrained encoder biases. & 
\textbf{Fig.\ref{fig:tokens_collapse_syn_all_exp_results}}: Visualization of token shrinkage on synthetic dataset. \par
\textbf{Fig.\ref{fig:token_shrinkage_and_pretrained_methods}}: Initialization induced shrinkage. &
Latent and reconstruction visualization. \\ 
\addlinespace[1.5ex]

C2 & Deferred Quantization reduces shrinkage and improves codebook coverage. & 
\textbf{Sec.~\ref{sec:deferred_quantization}}: Introduction of Deferred Quantization. \par \textbf{Fig.~\ref{fig:tokens_collapse_syn_all_exp_results} \& \textbf{Fig.~\ref{fig:token_collapse_real_data_results}}\& Tab.~\ref{tab:tab_all_tokenizer_imagenet} \& Tab.~\ref{tab:tab_tokenizer_odir}}: Deferred Quantization show better token spread and codebook utilization as well as improved reconstruction quality. & 
 r-FID $\downarrow$, LPIPS $\downarrow$, MSE $\downarrow$, Perplexity $\uparrow$, Codebook distance $\uparrow$ \\ 
\addlinespace[1.5ex]

C3 & Shrinkage is different from codebook collapse; fixing collapse alone is insufficient. & 
\textbf{Tab.~\ref{tab:tab_tokenizer_comparsion_with_simvq}}: Deferred Quantization enables gains where collapse-fixing fails. & 
r-FID $\downarrow$, LPIPS $\downarrow$, MSE $\downarrow$, Perplexity $\uparrow$, Codebook distance $\uparrow$ \\  
\addlinespace[1.5ex]

C4 & Generative diversity loss is caused by early quantization via latent support shrinkage, not discretization itself. & 
\textbf{Fig.\ref{fig:token_shrinkage_illustration}}: Illustration of token representation shrinkage limiting generative diversity. \par \textbf{Tab.\ref{tab:tab_generation_imagenet} and Tab. \ref{tab:tab_generation_odir}}: Quantified loss of diversity (ImageNet-100 \& ODIR). & 
g-FID $\downarrow$, Pixel Distance $\uparrow$, LPIPS Diversity $\uparrow$ \\ 
\addlinespace[1.5ex] % 在此调节行间距，1.5ex 约等于一个字符的高度

C5 & Mitigating shrinkage improves generation quality and diversity across models and datasets. & 
\textbf{Tab.~\ref{tab:tab_generation_imagenet} \& \ref{tab:tab_generation_odir}}: Consistent fidelity and diversity gains on ImageNet-100 and ODIR benchmarks. & 
g-FID $\downarrow$, LPIPS Diversity$\uparrow$, Pixel Distance $\uparrow$ \\ 

\bottomrule
\end{tabularx}
\end{table*}

It is important to note that in experiments involving generative models, GAN-based losses might introduce smoothing effects that could confound the analysis of token representation shrinkage. Therefore, in this section, we adopt VQ-VAE as the image tokenizer. The training loss includes \textit{codebook loss}, \textit{commitment loss}, \textit{MSE loss}, and \textit{perceptual loss}. Generative experimental results based on VQGAN are available in the appendix \ref{appendix_gan_results}.

% ------------ Real world data ----------- %
\begin{figure}[t]
    \centering
    \includegraphics[width=\linewidth ]{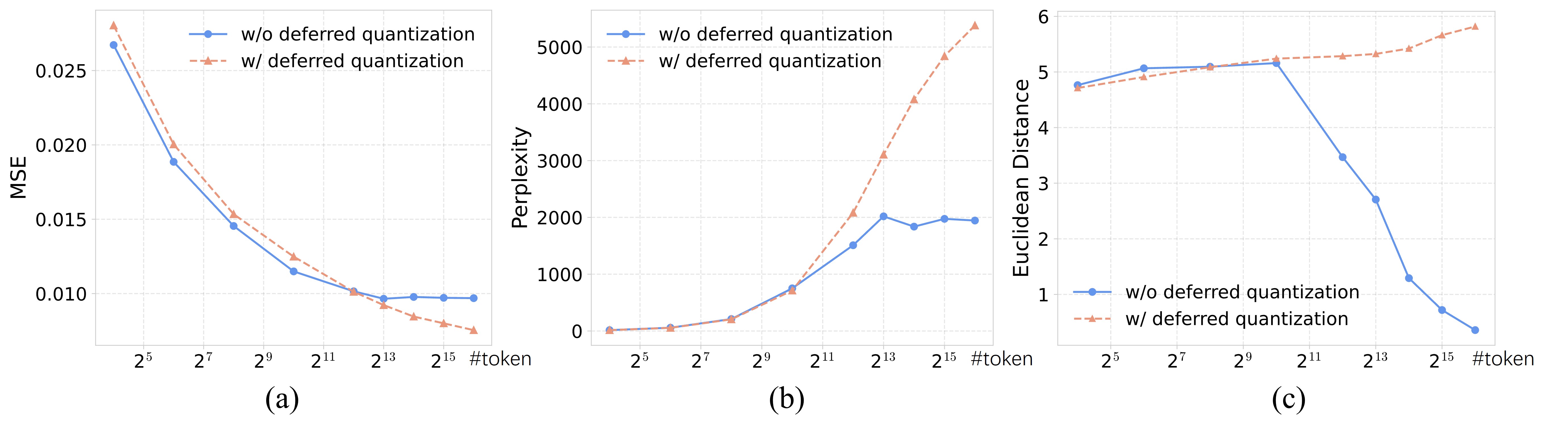}
    \caption{\textbf{Deferred Quantization alleviates the token representation shrinkage on CIFAR-10.}
(a) Token representation shrinkage in standard VQ impairs the model's ability to scale, leading to stagnating MSE. Deferred Quantization mitigates this shrinkage, allowing the model to achieve better reconstruction performance as number of token increase.
(b)Without Deferred Quantization, perplexity remains low, indicating highly uneven token usage. Deferred Quantization resolves this by spreading representations across the embedding space, ensuring high perplexity and efficient utilization of the available latent capacity. (c) The sharp drop in codebook Euclidean Distance for standard VQ indicates that tokens are clustering into a narrow region (shrinkage). Deferred Quantization maintains a high distance between tokens, preserving codebook coverage.
}
    \label{fig:token_collapse_real_data_results}
\end{figure}

\subsection{Setup}
\label{sec:experiment_setup}

\paragraph{Dataset}
As mentioned in Sec.~\ref{sec:Shrinkage Phenomena and Sythentic Experiments Results}, we conduct experiments on a synthetic dataset to validate our hypothesis regarding the causes of token representation shrinkage. The synthetic dataset consists of 10,000 data points, obtained by sampling 1,000 points from each of 10 Gaussian distributions with identical standard deviations but distinct means. This setup yields ten equally sized classes with similar distribution, designed to emphasize disproportionate token allocation and make token representation shrinkage patterns more easily observable. To investigate token representation shrinkage behavior under varying data complexity, we generate synthetic datasets with different input dimensionalities. To further validate the existence of token representation shrinkage, we additionally conduct experiments on CIFAR-10 and ImageNet-100.

For experiments on generative models, we adopt ImageNet-100, a subset of the ImageNet-1K dataset containing 100 classes. The original ImageNet-100 comprises approximately 130{,}000 training images and 5{,}000 test images. To better evaluate both reconstruction-FID (r-FID) and generation-FID (g-FID), we uniformly sample 20{,}000 images from all training classes to build the test dataset and construct an additional validation set containing 5{,}000 images.

For the medical domain, we adopt the Ocular Disease Recognition (ODIR) \citep{maranhao2020odir} dataset, which contains 6{,}716 fundus images labeled across 8 diagnostic categories. We use a 70\%/20\%/10\% split to partition the data into training, test, and validation sets.

\paragraph{Metrics}
For the synthetic dataset, we directly visualize the original data and its reconstructions, along with the corresponding token and embedding distributions, as shown in Fig.~\ref{fig:tokens_collapse_syn_all_exp_results}. For high-dimensional data, t-SNE is applied for dimensionality reduction prior to visualization.

To quantify token representation shrinkage, we report average Euclidean codebook distance and perplexity. The average pairwise Euclidean distance across the codebook serves as an indicator of token clustering, with lower values suggesting that code vectors have concentrated in a limited region. The perplexity, computed by the entropy over code usage likelihood, reflects the effective number of tokens being utilized and is maximized when all tokens are used uniformly.

To evaluate tokenizer reconstruction performance, we adopt reconstruction FID (r-FID), mean squared error (MSE), and LPIPS scores. For generative quality, we utilize generation FID (g-FID) as the primary metric. To assess diversity and distributional coverage of generated samples, we compute the average pairwise pixel-level distance between generated images. To reflect perceptual diversity, LPIPS Diversity is computed as the mean of learned perceptual distances between all pairs of generated images.

\paragraph{Training Configuration}
For generative model experiments, we follow the tokenizer framework proposed in VQGAN \citep{esser2021taming}. Due to resource limitation, we resize all input images to 128$\times$128 resolution and reduce the backbone’s channel size to 64. To preserve a 16$\times$16 latent spatial resolution, one downsampling layer and its upsampling layer are removed. All tokenizer experiments are conducted with a fixed codebook size of 16{,}384. To ensure feasibility under limited resources, we use the smallest generative model configurations. The MaskGIT generator employs a ViT \citep{dosovitskiy2020image} with depth of 24, while the VAR model uses a depth of 16.

\begin{table*}[t]
\caption{\textbf{Performance comparison of VAR tokenizers on the ImageNet-100 dataset.} SimVQ\citep{zhu2025addressing} is a well-known method to solve the code collapse problem.
The results show that mitigating code collapse alone does not lead to a substantial improvement, indicating that token representation shrinkage is different from code collapse.
}
\vspace{1.0em}
\centering
\renewcommand{\arraystretch}{1.2}
\setlength{\tabcolsep}{18pt}
\begin{tabular}{lccccc}
\toprule
VAR Tokenizer & r-FID ↓ &  MSE ↓ & LPIPS ↓ & Euc. ↑  & Perp. ↑ \\
\midrule
w/o Deferred & 5.39 & 2.60 & 1.85 & 1.19 & 2801.88  \\
\quad + SimVQ & 5.52 & 2.50 & 1.78 & 0.80 & 6920.23 \\
w/ Deferred & 5.04 & 2.22 & 1.63 & \textbf{7.56} & 7044.51 \\
\quad + SimVQ & \textbf{4.93} & \textbf{2.17} & \textbf{1.59} & 0.93 & \textbf{8222.83}  \\
\bottomrule
\\
\end{tabular}
\label{tab:tab_tokenizer_comparsion_with_simvq}
\vspace{-2.0em}
\end{table*}

\subsection{Analysis of Shrinkage in Tokenization}
\label{sec:exp_shrinkage_tokenization}

To validate that token representation shrinkage exists under real-world data conditions, we conduct experiments on CIFAR-10.
Additionally, we hypothesize that given a fixed dimensionality of the representation space, increasing the number of tokens tends to facilitate their clustering, thereby making token representation shrinkage more pronounced and more easily observable. Therefore, we evaluate VQ-VAE performance across varying token quantities.

As shown in Fig.~\ref{fig:token_collapse_real_data_results}, the original VQ model performs well when the number of tokens is relatively small. However, as the token count increases, particularly beyond $2^{12}$, its reconstruction performance deteriorates relative to the pretrained counterpart. Notably, the perplexity curve of the original VQ flattens after $2^{13}$ tokens, indicating poor token utilization. Additionally, its average Euclidean distance remains consistently lower than that of the pretrained model, suggesting a higher degree of similarity among tokens. These findings collectively indicate that shrinkage becomes increasingly severe as the token set grows, leading to reduced coverage of the representation space.

One possible reason why the model utilizing a pretrained AE backbone underperforms the original VQ-VAE at low token numbers is the gap between the continuous representations learned during pretraining and the discrete representations during finetuning, which poses challenges to the VQ learning process. However, this negative impact is outweighed by the benefits of mitigating token representation shrinkage as the codebook size increases. Overall, pretraining provides an effective intervention for mitigating shrinkage and shows promise in better exploiting large codebooks. It remains an open and valuable direction to explore how to mitigate the performance gap when the number of tokens is small.

We further analyze token representation shrinkage under the multi-scale tokenizer of VAR on ImageNet-100. To highlight that token representation shrinkage is distinct from code collapse, we incorporate SimVQ \cite{zhu2025addressing}, an efficient method specifically designed to mitigate code collapse. The results are reported in Tab.~\ref{tab:tab_tokenizer_comparsion_with_simvq}. We observe that when token representation shrinkage is present, addressing only code collapse with SimVQ does not yield a substantial improvement in tokenizer performance. In contrast, when both token representation shrinkage and code collapse are simultaneously resolved, the tokenizer achieves a notable performance gain. This supports the claim that shrinkage is a separate bottleneck beyond dead-token (code-collapse) issue.

\subsection{Analysis of Shrinkage in Generation}
\label{sec:exp_shrinkage_generation}

\subsubsection{ImageNet-100}

\paragraph{Tokenizer performance.}
Both types of original tokenizers exhibit clear token representation shrinkage, as shown in Tab.~\ref{tab:tab_all_tokenizer_imagenet}. For the tokenizer used in MaskGIT \citep{chang2022maskgit}, we observe limited variation among tokens indicated by relatively small Euclidean distances (6.45 \textit{vs.} 18.75), reflecting high similarity between tokens. In addition, the tokenizer exhibits low perplexity (924.57 \textit{vs.} 5311.88), suggesting that only a small subset of tokens is frequently utilized. Together, these observations imply that token usage is poorly aligned with the embedding space, pointing to token representation shrinkage.

After pretraining, tokens are more evenly utilized and better aligned with the embedding space, indicating that the intervention mitigates token representation shrinkage. As a result, the pretrained tokenizer achieves improved reconstruction performance, with lower r-FID (8.58 \textit{vs.} 12.22), LPIPS (2.34 \textit{vs.} 2.70), and MSE (3.28 \textit{vs.} 3.91). A similar pattern is observed for the multi-scale tokenizer in VAR: without pretraining, severe shrinkage is evident, whereas pretraining alleviates it and improves reconstruction.

\begin{table*}[t]

\caption{\textbf{Performance evaluation of various tokenizer on the ImageNet-100 dataset.}
\ding{51} indicates that Deferred Quantization is applied, whereas \ding{55} indicates that Deferred Quantization is not applied.
}
\vspace{1.0em}
\label{tab:tab_all_tokenizer_imagenet}
\centering
\renewcommand{\arraystretch}{1.0}
\setlength{\tabcolsep}{10pt}
\begin{tabular}{lcccccc}
\toprule
Tokenizer & w/ Deferred? & r-FID ↓ &  MSE ↓ & LPIPS ↓ & Euc. ↑  & Perp. ↑ \\
\midrule
\multirow{2}{*}{MaskGIT}
~ & \ding{51}  & \textbf{8.58} & \textbf{3.28} & \textbf{2.34} & \textbf{18.75} & \textbf{5311.88} \\
~ & \ding{55} & 12.22 & 3.91 & 2.70 & 6.45 & 924.57 \\
\midrule
\multirow{2}{*}{VAR}
~  & \ding{51}  & \textbf{5.04} & \textbf{2.22} & \textbf{1.19} & \textbf{7.56} & \textbf{7044.51} \\
~ & \ding{55} & 5.39 & 2.60 & 1.85 & 1.19 & 2801.88  \\
\bottomrule
\\
\end{tabular}
\vspace{-2.0em}
\end{table*}

\begin{table}[t]
\centering
\caption{\textbf{Comparison of generative models in the presence and absence of token representation shrinkage on ImageNet-100.} \ding{51} indicates that token representation shrinkage is mitigated, whereas \ding{55} indicates without mitigation. Token representation shrinkage impairs both generative quality (g-FID) and diversity (pixel distance and LPIPS diversity).}
\label{tab:tab_generation_imagenet}
\vspace{1.0em}
\renewcommand{\arraystretch}{1.1}
%\resizebox{\columnwidth}{!}{%
\begin{tabular}{lcccc}
\toprule
Model & w/ Deferred? & g-FID ↓ & Pixel Dist. ↑ & LPIPS D. ↑\\
\midrule
\multirow{2}{*}{MaskGIT}
~ & \ding{51} & \textbf{14.60} & \textbf{80.77} & \textbf{0.677} \\
~ & \ding{55} & 14.75         & 75.89          & 0.668 \\
\midrule
\multirow{2}{*}{VAR}
~ & \ding{51} & \textbf{10.44} & \textbf{75.75} & \textbf{0.667} \\
~ & \ding{55} & 12.88          & 70.69          & 0.633 \\
\bottomrule
\end{tabular}%
% }
\end{table}

\paragraph{Generative performance.}
Token representation shrinkage degrades downstream generation, manifesting as both worse sample quality and reduced diversity (often as mode collapse), as shown in Tab.~\ref{tab:tab_generation_imagenet}. For MaskGIT, shrinkage leads to worse g-FID and reduced pairwise pixel distance and LPIPS diversity among generated samples, indicating that outputs become more similar and cover fewer variations. This aligns with the mechanism that shrinkage narrows the token distribution available to the generator.

For VAR, we observe a consistent pattern: shrinkage in its multi-scale tokenizer leads to reduced generation quality and a drop in diversity. These results reinforce the conclusion that insufficient token coverage limits the model’s ability to represent the full data distribution, which propagates to mode collapse / reduced diversity in downstream generation. Qualitative examples are shown in \ref{sec_visual_generated_examples}.

\subsubsection{ODIR Medical Dataset}

To further validate our findings, we conduct experiments on the ODIR medical image dataset. For VAR, we again confirm the presence of token representation shrinkage, as shown in Tab.~\ref{tab:tab_tokenizer_odir}. We also observe a corresponding decline in generative performance, including reductions in both sample quality and diversity (Tab.~\ref{tab:tab_generation_odir}), consistent with our observations on natural image datasets.

\begin{table*}[t]
\caption{\textbf{Performance evaluation of VAR tokenizer on the ODIR medical dataset.}
\ding{51} indicates that token representation shrinkage is mitigated, whereas \ding{55} indicates without mitigation. The results indicate that token representation shrinkage also exists in a real-world medical setting.
}
\vspace{1.0em}
\centering
\renewcommand{\arraystretch}{1.0}
\setlength{\tabcolsep}{9pt}
\begin{tabular}{lc ccccc}
\toprule
Model & w/ Deferred? & r-FID ↓ &  MSE ↓ & LPIPS ↓ & Euc. ↑  & Perp. ↑\\
\midrule
\multirow{2}{*}{VAR}
~ & \ding{51} & 11.04 & \textbf{2.05} & \textbf{6.79} & \textbf{8.50} & \textbf{5396.17}\\
~ & \ding{55} & \textbf{10.91} & 2.57 & 8.79 & 1.27 & 940.55 \\
\bottomrule
\end{tabular}
\label{tab:tab_tokenizer_odir}
\end{table*}

\begin{table}[t]
\centering
\caption{\textbf{Generative results of VAR on the ODIR medical dataset.}
\ding{51} indicates that De\-ferred Quantization is applied, whereas \ding{55} indicates that De\-ferred Quantization is not applied. The results indicate that token representation shrinkage also exists in a real-world medical setting.
}
\vspace{1.0em}
\label{tab:tab_generation_odir}
\renewcommand{\arraystretch}{1.2}
\begin{tabular}{lcccc}
\toprule
Model &w/ Deferred?& g-FID ↓ & Pixel Dist. ↑ & LPIPS D. ↑\\
\midrule
\multirow{2}{*}{VAR}
~ & \ding{51} & \textbf{34.33} & \textbf{49.83} & \textbf{0.401} \\
~ & \ding{55} & 37.65          & 49.01          & 0.390 \\
\bottomrule
\end{tabular}%

\end{table}

We observe that r-FID slightly increases for the VAR tokenizer on ODIR when shrinkage is absent. We attribute this to the relatively small size of the ODIR test set (about 1,200 images), which can introduce variability in FID-based evaluations. As shown in Tab.~\ref{tab:tab_tokenizer_odir}, other quality metrics such as MSE and LPIPS consistently improve after applying our pretraining intervention, indicating clear gains in reconstruction quality. Another contributing factor is that FID relies on features extracted by an Inception-V3 network pretrained on ImageNet-1K, which may not capture semantic differences reliably for domain-specific medical images such as fundus photographs.

However, results for MaskGIT on ODIR do not fully align with expectations. Although the tokenizer shows clear evidence of token representation shrinkage, generated images do not exhibit a noticeable loss of diversity. One possible explanation is the relatively small size of the ODIR dataset (approximately 6,000 images) together with its limited variability, as all classes consist of closely related ocular images. Such characteristics may reduce the sensitivity of downstream generative evaluation and mask the impact of shrinkage. Detailed quantitative analyses are included in the supplementary material.

\section{Conclusion}

In this work, we systematically analyze \emph{token representation shrinkage} in vector quantization as a critical and previously underappreciated factor that can drive reduced diversity and mode collapse in token-based generative models. We show that a commonly adopted codebook initialization strategy---initializing tokens from an untrained encoder---induces a geometric shrinkage in the token embedding space, leading to poorly spread tokens, unbalanced token usage, and insufficient latent support for downstream generation.

To convert this diagnosis into an actionable fix, we propose \emph{De\-ferred Quantization}, a simple two-stage training protocol that solve the token representation shrinkage problem. Across theoretical analysis and extensive experiments on synthetic data, natural images, and medical images, we find that applying De\-ferred Quantization consistently expands token coverage, improves tokenizer reconstruction quality, and propagates to improved sample diversity and reduced mode collapse in downstream generators. Overall, our results highlight that tokenizer design should go beyond reconstruction-centric criteria, and that controlling early-stage quantization dynamics is a simple yet effective lever for building diversity-preserving discrete representations.

\bibliography{main}
\bibliographystyle{tmlr}

\appendix
\section{Appendix}

\subsection{Implementation Details}
\label{appendix_Implementation_details}

\paragraph{Sythentic Dataset}
Our synthetic dataset includes 10 clusters, each with 1,000 data points sampled from a Gaussian distribution and standardized using Standard Scaler. Our VQ-VAE comprises an MLP-based encoder/decoder with three linear layers and uses the ReLU activation function. Training is facilitated by the AdamW optimizer, with a learning rate of 0.001. Additional specifications include a codebook size of 128, a hidden dimension of 32, a batch size of 256, a beta of 0.25, and a decay rate ($\gamma$ for EMA) of 0.9. In experiments, the autoencoder is trained for 100 epochs. The fine-tuned VQ-VAE and the original VQ-VAE are trained for 100 and 200 epochs respectively.

\paragraph{CIFAR-10}
For CIFAR-10, our VQ-VAE adopts downsampling using a CNN with a downsample channel of 128, and the model includes two residual blocks with a hidden channel size of 64. The codebook size is set at 512 with a token dimension of 64. The learning rate is 3e-4, using the Adam optimizer with amsgrad set to true. The beta is 0.25 and the decay rate is 0.99. The codebook size in experiments is varied from 16 to 65,536, with embedding and token adopting the size of 32. And we pretrain AE for 150 epochs and fine-tune the VQ-VAE for 150 epochs.

\paragraph{ImageNet-100 \& ODIR}
As described in the main text, we modified the tokenizer by removing one downsampling layer along with its corresponding upsampling layer and reducing the backbone’s channel size to 64. We employed both ReduceLROnPlateau and Cosine annealing schedulers to train the tokenizer. Detailed configurations can be found in the provided codebase (\textit{config.yaml}). The initial learning rate was set to $1 \times 10^{-4}$, with a minimum of $1 \times 10^{-6}$, and early stopping was applied. For ImageNet-100, we trained selected the checkpoint with the best FID on the validation set for downstream tasks on ImageNet-100, while the final checkpoint was used for the ODIR dataset. VAR and MaskGIT were trained with ReduceLROnPlateau scheduling and early stopping.

It is important to note that when the codebook size is large, it is infeasible to initialize it using embeddings from a single batch. Therefore, on CIFAR-10, ImageNet-100, and ODIR, we initialize the codebook using embeddings collected from multiple batches. Specifically, we maintain an embedding-to-token ratio of 10:1 on CIFAR-10, and 2:1 on ImageNet-100 and ODIR. Moreover, to improve computational efficiency, we compute LPIPS diversity using a 5,000-image subset of the ImageNet-100 test set. For the ODIR dataset, we use the full original test set.

 All tokenizers and generative models are trained on 2 A100 GPUs with 40 GB memory. Training the tokenizers on ImageNet-100 typically takes 1.5 to 3 days, while training the generative models requires 3--6 days depending on the setting.

 \paragraph{Initialization Strategy} For codebook initialization, a widely used initialization strategy  is K-means\citep{zeghidour2021soundstream}.
It uses the encoder output $\mathcal{Z}$ 
and perform K-means algorithm to initialize the tokens $\mathcal{T}$, where $N$ is the number of encoder output and $S$ is the number of tokens.
The initialization aims to minimize the total distance from each vector $z_j$ to its nearest token $t_k$. The optimizing function is shown in equation \ref{K_means_initialization_equ},
\begin{equation}
    \min \sum_{j=1}^N\sum_{k=1}^S r_{jk} \|z_j - t_k\|^2,\label{K_means_initialization_equ}
\end{equation}
where $r_{jk} = 1$ if $z_j$ is assigned to cluster center $t_k$, otherwise $r_{jk} = 0$  .

\begin{table}[ht] 
\centering
\footnotesize
\renewcommand{\arraystretch}{1.0} 
\setlength{\tabcolsep}{4pt}
\caption{\textbf{Comparison of generative models based on VQGAN on ImageNet-100.} \ding{51} indicates that token representation shrinkage is mitigated, whereas \ding{55} indicates without mitigation.}
\vspace{1.0em}
\label{tab:tab_gan_generation_imagenet}
\begin{tabular}{lcccc}
\toprule
Model & w/ Deferred? & g-FID ↓ & Pixel Dist. ↑ & LPIPS D.↑ \\
\midrule
\multirow{2}{*}{MaskGIT} 
& \ding{51}  & \textbf{10.85} & \textbf{79.17} & 0.712 \\
& \ding{55}  & 12.25 & 78.83 & \textbf{0.714} \\
\midrule
\multirow{2}{*}{VAR} 
& \ding{51}  & 8.30 & \textbf{77.10} & \textbf{0.719} \\
& \ding{55}  & \textbf{7.83} & 73.37 & 0.714 \\
\bottomrule
\end{tabular}
\end{table}

\subsection{Visualization of Embeddings}

\begin{figure*}[t]
    \centering
    \includegraphics[width=0.8\linewidth]{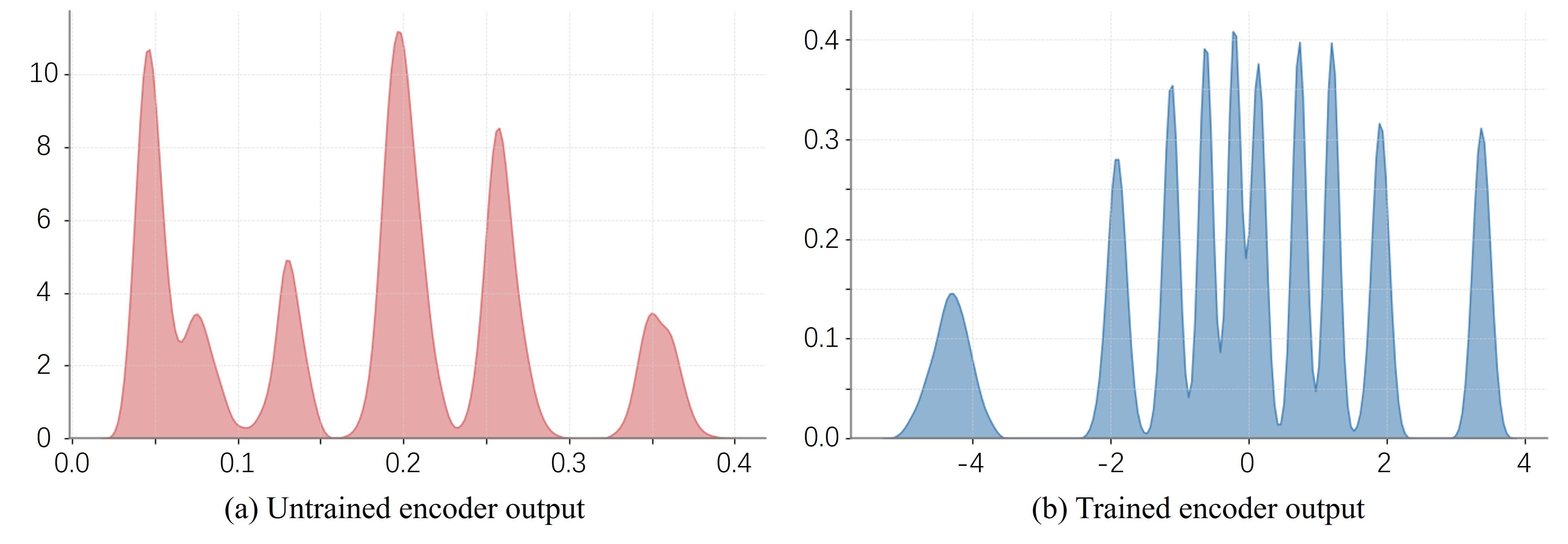}
    \caption{ \textbf{Distribution of untrained and trained encoder's output on synthetic dataset.} (a) Untrained encoder's output has fewer peaks than 10 peaks of input and shrinks into a relatively small range. (b) Trained encoder's output displays 10 peaks, which is the same as the input. This phenomena inspire us that pretraining might be an invention for token representation shrinkage.}
\label{fig:token_shrinkage_encoder_output}
\end{figure*}

% comparison between untrained and trained encoder's output embeddings
In our synthetic experiments, we find that the outputs of an untrained encoder are predominantly concentrated between $0.1$ and $0.4$ (Fig.\ref{fig:token_shrinkage_encoder_output} a) and exhibit only 6 distinct peaks, despite the input dataset containing 10 modes. This phenomenon motivates our pretraining intervention, which is designed to investigate the problem of token representation shrinkage.

\subsection{Visualization of generated examples}
\label{sec_visual_generated_examples}
\begin{figure*}[t]
    \centering
    \includegraphics[width=\linewidth]{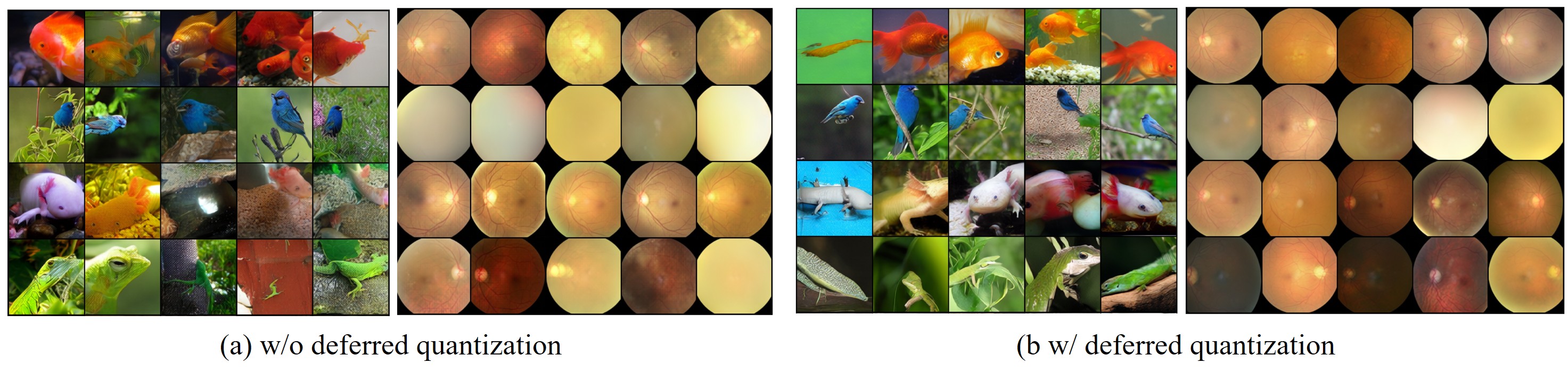}
    \caption{\textbf{Images generated using VAR.}
(a) ImageNet (a.left) and real-world medical images of eyes (a.right) generated by VAR \textbf{w/o deferred quantization}.
(b) ImageNet (b.left) and real-world medical images of eyes (b.right) generated by VAR \textbf{w/ deferred quantization}.}

\label{fig:token_shrinkage_generated_image_all}
\end{figure*}

As mentioned in previous section, for VAR generative model, we observe a consistent pattern: shrinkage in its multi-scale tokenizer leads to reduced generation quality and a drop in diversity. These results reinforce the conclusion that insufficient token coverage limits the model’s ability to represent the full data distribution, which propagates to mode collapse / reduced diversity in downstream generation. Qualitative examples are shown in Fig.~\ref{fig:token_shrinkage_generated_image_all}.

\subsection{VQGAN based Generation Results}
\label{appendix_gan_results}

\begin{table}[ht] 
\centering
\footnotesize % 
\renewcommand{\arraystretch}{0.95} 
\setlength{\tabcolsep}{6pt} 
\caption{\textbf{ODIR generation based on VQGAN of VAR.} \ding{51} indicates that token representation shrinkage is mitigated, whereas \ding{55} indicates without mitigation. Token representation shrinkage impairs generative quality and diversity.}
\vspace{1.0em}
\label{tab:tab_gan_generation_odir}
\begin{tabular}{lcccc}
\toprule
Model & w/ Deferred? & g-FID ↓ & Pixel Dist. ↑ & LPIPS D.↑ \\
\midrule
\multirow{2}{*}{VAR} 
& \ding{51} & 29.65 & \textbf{50.56} & 0.550 \\
& \ding{55} & \textbf{27.89} & 45.14 & \textbf{0.570} \\
\bottomrule
\end{tabular}
\end{table}

%%%%%%%%%%%%%%%%%%%%%%%%%%% Tokenizer VQGAN %%%%%%%%%%%%%%%%%%%%%%%%%%%%%%%%
\begin{table*}[!t]
\caption{\textbf{Evaluation of tokenizers trained with GAN loss on the ImageNet-100 dataset.} 
\ding{51} indicates that token representation shrinkage is mitigated, whereas \ding{55} indicates without mitigation. MSE and LPIPS values in the table are scaled. To recover the actual values, multiply them by $10^{-4}$ and $10^{-3}$, respectively.
}
\vspace{1.0em}
\centering
\renewcommand{\arraystretch}{0.95}
\setlength{\tabcolsep}{10pt}
\begin{tabular}{lcccccc}
\toprule
Tokenizer & w/ Deferred? & r-FID ↓ &  MSE ↓ & LPIPS ↓ & Euc. ↑  & Perp. ↑ \\
\midrule
\multirow{2}{*}{MaskGIT}
~  & \ding{51}  & \textbf{5.28} & \textbf{3.96} & \textbf{2.40} & \textbf{18.75} & \textbf{5575.92} 
\\
~ & \ding{55} & 6.40 & 4.58 & 2.70 & 6.52 & 920.95  
\\
\midrule
\multirow{2}{*}{VAR}
~  & \ding{51}  & 2.13 & \textbf{2.59} & \textbf{1.73} & \textbf{7.50} & \textbf{7143.41} 
\\
~ & \ding{55} & \textbf{2.09} & 3.00 & 1.87 & 1.13 & 2814.91  
\\
\bottomrule
\end{tabular}
\label{tab:tab_tokenizer_vqgan_imagenet}
\end{table*}

\vspace{1.0em}

%%%%%%%%%%%%%%%%%%%%%%%%%%% Tokenizer VQGAN %%%%%%%%%%%%%%%%%%%%%%%%%%%%%%%%
\begin{table*}[!t]
\caption{\textbf{Evaluation of VAR tokenizers trained with GAN loss on the ODIR dataset.} \ding{51} indicates that token representation shrinkage is mitigated, whereas \ding{55} indicates without mitigation. To recover the actual MSE and LPIPS, multiply them by $10^{-5}$ and $10^{-4}$, respectively. 
}
\vspace{1.0em}
\centering
\renewcommand{\arraystretch}{1.0}
\setlength{\tabcolsep}{10pt}
\begin{tabular}{lcccccc}
\toprule
Tokenizer & w/ Deferred? & r-FID ↓ &  MSE ↓ & LPIPS ↓ & Euc. ↑  & Perp. ↑
\\
\midrule
\multirow{2}{*}{VAR}
~ & \ding{51}  & \textbf{6.90} & \textbf{2.19} & \textbf{7.15} & \textbf{8.50} & \textbf{5451.76}
\\
~ & \ding{55} & 8.12 & 2.66 & 9.39 & 1.27 & 801.07
\\
\bottomrule
\\
\end{tabular}
\label{tab:tab_tokenizer_vqgan_odir}
\end{table*}

\vspace{-1.0em}

\begin{table*}[!t]
\caption{\textbf{MaskGIT tokenizer and generative performance on medical dataset.} 
$\mathcal{L}_{\text{GAN}}$ denotes whether GAN loss was used during training. 
\ding{51} indicates that token representation shrinkage is mitigated, whereas \ding{55} indicates without mitigation. 
To recover the actual MSE and LPIPS values, multiply them by $10^{-5}$ and $10^{-4}$.}
\vspace{1.0em}
\label{tab:tab_odir_maskgit}
\centering
{\small
\renewcommand{\arraystretch}{1.0}
\setlength{\tabcolsep}{6pt}
\resizebox{0.9\textwidth}{!}{%
\begin{tabular}{cccccccccc}
\toprule
$\mathcal{L}_{\text{GAN}}$ & w/ Deferred? & r-FID ↓ & MSE ↓ & LPIPS ↓ &  Euc. ↑ & Perp. ↑ & g-FID ↓ & Pixel Dist. ↑ &  LPIPS D. ↑ \\
\midrule
\multirow{2}{*}{No}
& \ding{51} & \textbf{10.33} & \textbf{3.16} & \textbf{0.90} & \textbf{12.70} & \textbf{3346.04} & 38.04 & 48.44 & 0.370\\
& \ding{55} & 11.96 & 4.19 & 1.16 & 4.92 & 438.36 & \textbf{27.95} & \textbf{49.73} & \textbf{0.397}\\
\midrule
\multirow{2}{*}{Yes}
& \ding{51} & \textbf{10.47} & \textbf{3.67} & \textbf{1.02} & \textbf{12.60} & \textbf{3211.72} & 38.30 & 48.15 & 0.371 \\
& \ding{55} & 12.72 & 4.64 & 1.30 & 4.93 & 432.07 & \textbf{32.60} & \textbf{49.28} & \textbf{0.402} \\
\bottomrule
\end{tabular}
}
}
\end{table*}

As shown in Tab.\ref{tab:tab_tokenizer_vqgan_imagenet} and Tab.\ref{tab:tab_tokenizer_vqgan_odir}, low Euclidean distance and low perplexity indicate the presence of token representation shrinkage in VQGAN. And token representation shrinkage still degrades reconstruction quality. For generative models, Tab.\ref{tab:tab_gan_generation_imagenet} and Tab.\ref{tab:tab_gan_generation_odir} demonstrate the impact of token representation shrinkage might affect generative diversity and fidelity. In particular, Pixel Distance is consistently lower under shrinkage, reflecting reduced pixel diversity in generated samples. However, we found better r-FID and g-FID scores in the presence of shrinkage. We hypothesize that this is due to GAN loss might enhance the decoder’s expressive capacity, which enables it to compensate for the effects of token representation shrinkage.


\subsection{Results for MaskGIT on Medical dataset}
As shown in Tab. \ref{tab:tab_odir_maskgit}, when token representation shrinkage occurs, the tokenizer in MaskGIT also exhibits a decline in reconstruction quality on the ODIR dataset. However, the generative model unexpectedly displays high quality and diversity under token representation shrinkage. We hypothesize that this counterintuitive result may stem from the limited data available for training and evaluation. The ODIR dataset contains only about 6,000 images, all of which are restricted to ocular imagery, resulting in lower diversity.

\end{document}